\documentclass[letterpaper]{article} 
\usepackage[submission]{aaai23}  
\usepackage{times}  
\usepackage{helvet}  
\usepackage{courier}  
\usepackage[hyphens]{url}  
\usepackage{graphicx} 
\usepackage{natbib}  
\usepackage{caption} 
\usepackage{algorithm}
\usepackage{algorithmic}
\usepackage{amsmath,amsfonts,bm}
\usepackage{newfloat}
\usepackage{listings}
\DeclareCaptionStyle{ruled}{labelfont=normalfont,labelsep=colon,strut=off} 
\floatstyle{ruled}
\newfloat{listing}{tb}{lst}{}
\floatname{listing}{Listing}
\title{Large scale global optimization of ultra-high dimensional non-convex landscapes based on generative neural networks}
\author {
    Jiaqi Jiang,\textsuperscript{\rm 1}
    Jonathan A. Fan, \textsuperscript{\rm 1}
}
\affiliations {
    \textsuperscript{\rm 1} Stanford University, Department of Electrical Engineering, Stanford, CA, United States\\
    jiangjq@stanford.edu, jonfan@stanford.edu
}

\begin{document}

\maketitle

\begin{abstract}
We present a non-convex optimization algorithm metaheuristic, based on the training of a deep generative network, which enables effective searching within continuous, ultra-high dimensional landscapes.  During network training, populations of sampled local gradients are utilized within a customized loss function to evolve the network output distribution function towards one peaked at high performing optima.  The deep network architecture is tailored to support progressive growth over the course of training, which allows the algorithm to  manage the curse of dimensionality characteristic of high dimensional landscapes.  We apply our concept to a range of standard optimization problems with dimensions as high as one thousand and show that our method performs better with fewer functional evaluations compared to state-of-the-art algorithm benchmarks.  We also discuss the role of deep network over-parameterization, loss function engineering, and proper network architecture selection in optimization, and why the required batch size of sampled local gradients is independent of problem dimension.  These concepts form the foundation for a new class of algorithms that utilize customizable and expressive deep generative networks to solve non-convex optimization problems.

\end{abstract}

\section{Introduction}


High dimensional, non-convex optimization problems are pervasive in many scientific and engineering domains, including computational materials science \cite{chen2017sgo, jorgensen2018exploration}, electromagnetics \cite{piggott2015inverse, sell2017large}, circuits design \cite{held2011combinatorial},  process engineering \cite{sieniutycz2009energy}, and systems biology \cite{balsa2012global}.  These problems are known to be very difficult to solve because they are NP-hard, and algorithms aiming to definitively search for the global optimum, such as branch and bound methods, cannot practically scale to high dimensional systems.  As such, various algorithm heuristics have been developed, ranging from evolutionary metaheuristics to Bayesian optimization \cite{pelikan1999boa, snoek2015scalable}, which use judicious sampling of the landscape to identify high performing optima.  In all cases, it remains challenging to apply these algorithms to ultra-high dimensional spaces with dimensions of hundreds to thousands due to the curse of dimensionality. 


The explosion of interest and research in deep neural networks over the last decade has presented new opportunities in optimization, as the process of training a deep network involves solving a high dimensional optimization problem.  To this end, gradient-based optimization metaheuristics termed global topology optimization networks (GLOnets) \cite{jiang2020simulator, jiang2019global} were recently proposed that use the training of a deep generative network to perform non-convex optimization. The concept applies to optimization problems where $\mathbf{x}$ is a $d$-dimensional variable and the goal is to maximize the smoothly varying, non-convex objective function $f(\mathbf{x})$.  To run the metaheuristic, the generative network is first initialized so that it outputs a distribution of $\mathbf{x}$ values that spans the full optimization landscape.  Over the course of network training, this distribution is sampled, $f(\mathbf{x})$ and local gradients are computed for these sampled points, and these values are incorporated into a customized loss function and backpropagated to evolve and narrow the distribution around high performing optima.  Initial demonstrations indicate that GLOnets can perform better than standard gradient-based optimizers and global search heuristics for various non-convex optimization problems.  However it is unable to extend to high dimensional problems in its current form, and the lack of interpretability with this black box algorithm has made it difficult to understand if and how it can to adapt to more general problems, including high dimensional problems.


In this Article, we introduce the progressive growing GLOnet (PG-GLOnet) in which optimization within an ultra-high dimensional non-convex landscape is mediated through the training of a progressive growing deep generative network.  Our tailoring of the network architecture for this optimization task serves to incorporate knowledge and assumptions about the optimization landscape into the metaheuristic, which is a requirement for tractably navigating ultra-high dimensional landscapes.  We also explain how the algorithm works to smoothen the design landscape, how evaluation of the loss function serves as a gradient estimation calculation, and why the number of required functional evaluations is independent of problem dimension.  With standard benchmarking test functions, we show that our concept performs better than state-of-the-art algorithms with fewer functional evaluations for one thousand dimensional problems.  We anticipate that the customization of network architectures within the GLOnets framework will seed new connections between deep learning and optimization.

\section{Progressive Growing GLOnets Algorithm and Benchmarking}

\begin{figure*}[ht]
    \centering
    \includegraphics[width=15cm]{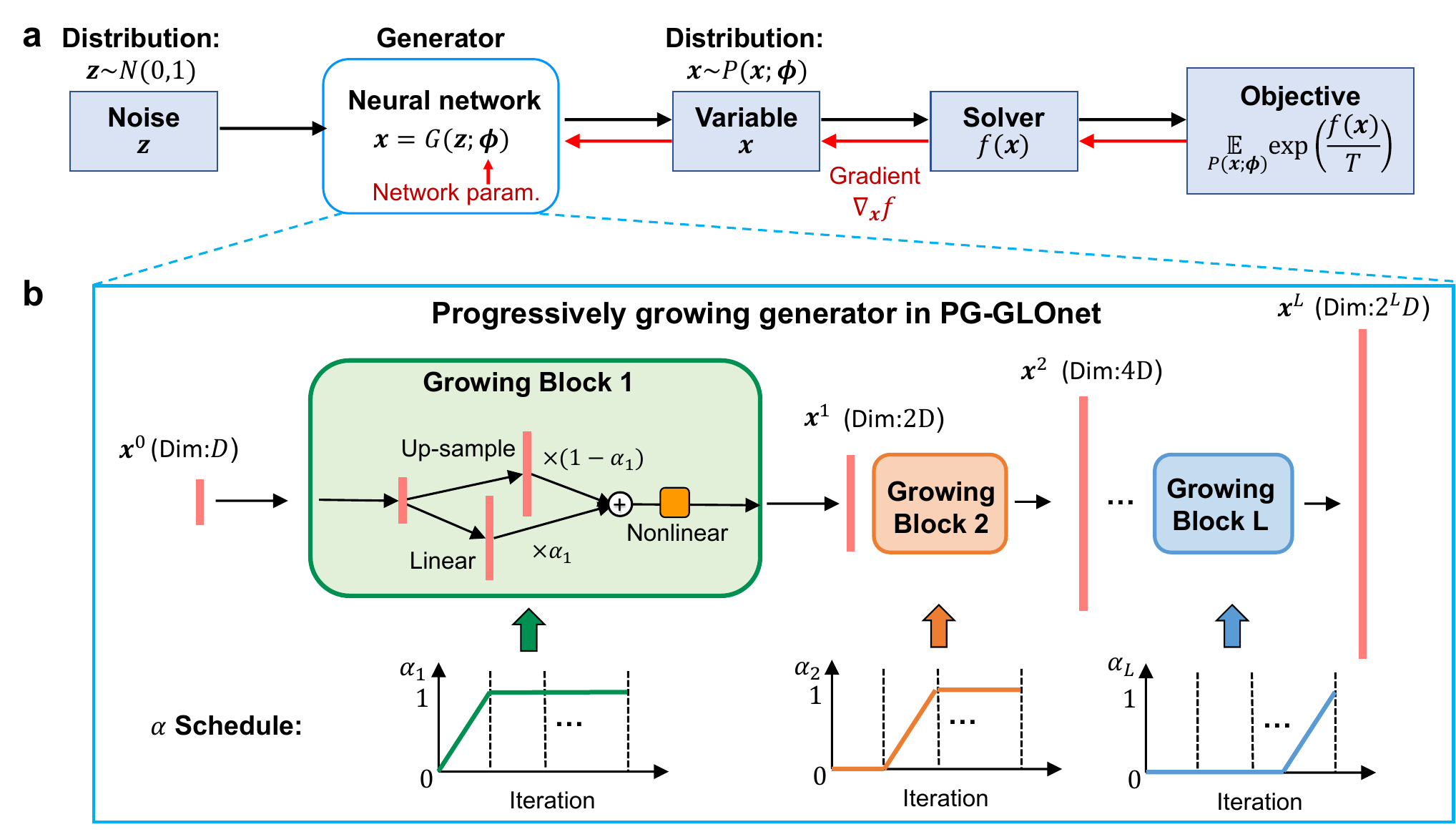}
    \caption{(a) Framework of the GLOnet algorithm. A deep generative network produces a distribution of design variables $\mathbf{x}$ and the distribution is narrowed around high performing optima by backpropagation. (b) PG-GLOnet generator architecture.  A set of growing blocks are implemented and activated over the course of network training to enable optimization in very high-dimensional landscapes.}
    \label{fig:1}
\end{figure*}


The PG-GLOnet concept builds on the foundation of the original GLOnet algorithm, which we briefly review here.  The optimization problem to be solved with GLOnets can be written in the following form:
\begin{equation}
    \max_{\mathbf{x}} f(\mathbf{x})
    \label{eq:1}
\end{equation}
where $f(\mathbf{x})$ is a non-convex, continuous objective function with feasible gradients.  With GLOnets, this optimization problem is indirectly solved through the training of a general neural network (Figure \ref{fig:1}a), where the input is a $d$-dimensional random variable $\mathbf{z}$ with a standard normal distribution and the output is a distribution of $\mathbf{x}$'s.  The generator therefore serves to map $\mathbf{z}$ onto $\mathbf{x} = G(\mathbf{z}; \phi)$ with a distribution $P(\mathbf{x}; \phi)$, where $\mathbf{\phi}$ denotes the trainable neural network parameters. The optimization objective for the generator is defined as:
\begin{equation}
    L = \max_{\mathbf{\phi}} \mathop{\mathbb{E}}_{\mathbf{x} \sim P(\mathbf{x}; \phi)} \exp\left[ \frac{f(\mathbf{x})}{T} \right]
    \label{eq:2}
\end{equation}
The distribution that maximizes this expected value is a delta function centered at the global optimum, and as such, an ideally trained generator will produce a narrow distribution centered at the global optimum, thereby solving the original optimization problem.  The use of the exponential function and the hyperparameter $T$ in the optimization objective further enhance the valuation of the global optimum, and more generally high performing optima, in the design space.

Generator training is consistent with conventional deep learning training methods: gradients of the objective function with respect to network parameters, $\nabla_{\phi}\mathbb{E}f$, are calculated through backpropagation, and they are used to iteratively optimize $\mathbf{\phi}$ using standard gradient-based methods. In practice, the objective function is approximated by a batch of $M$ samples. $P(\mathbf{x}; \phi)$, on the other hand, is typically implicit and cannot be directly sampled.  To circumvent this issue, we draw $M$ samples $\{\mathbf{z}^{(m)}\}_{m=1}^M$ from the standard normal distribution, transform them to $\{\mathbf{x}^{(m)}\}_{m=1}^M$, and then approximate $L$ and its gradient $\nabla_\phi L$ with respect to network parameters $\phi$:
\begin{equation}
     L \approx \frac{1}{M} \sum_{m=1}^{M} \exp\left[ \frac{f(\mathbf{x}^{(m)})}{T} \right] 
     \label{eq:3}
\end{equation}
\begin{equation}
    \nabla_\phi L \approx \frac{1}{M} \sum_{m=1}^{M} \frac{1}{T}\exp\left[ \frac{f(\mathbf{x}^{(m)})}{T} \right] \nabla_{\mathbf{x}}f \cdot D_{\phi}\mathbf{x}^{(m)}
    \label{eq:4}
\end{equation}
$\nabla_{\mathbf{x}}f = [\frac{\partial f}{\partial x_1}, \frac{\partial f}{\partial x_2}, \dots, \frac{\partial f}{\partial x_d}]$ are the gradients of $f(\mathbf{x})$ and $D_{\phi}\mathbf{x} = \frac{\partial (x_1, x_2, \dots)}{\partial(\phi_1, \phi_2, ...)}$ is the Jacobian matrix. Evaluation of $f(\mathbf{x})$ is usually performed by a numerical simulator and the gradient of $f(\mathbf{x})$ can be calculated explicitly or by auto-differentiation for analytic expressions, or by the adjoint variables method (AVM).  


In the initial conception of GLOnet, which we term FC-GLOnet, the generative network was a fully connected deep network and was capable of effectively addressing optimization problems with a modest number of dimensions.  However, it was found to be ineffective at optimizing within very high dimensional landscapes due to the curse of dimensionality, which makes a direct search for the global optimum within a full, high dimensional landscape an intractable proposition.  We therefore propose the PG-GLOnet, which utilizes a generative network that outputs a distribution that gradually grows from a coarse, low dimensional space to a fine, high dimensional space. By tailoring the network architecture in this way, we regularize the optimization process to take place over differing degrees of optimization landscape smoothing, enabling our search process to be computationally efficient and tractable.






The PG-GLOnet generator architecture is shown in Figure \ref{fig:1}b. The progressive growth concept is inspired by progressively growing GANs \cite{karras2017progressive} that have been developed in the computer vision community to process images with increasing spatial resolution during network training.  The input to the network is a $D$-dimensional random vector $\mathbf{x}^0$, and its dimension is much smaller than that of $\mathbf{x}$.  With $L$ growing blocks, the network simultaneously transforms and increases the dimensionality of the input vector, and its output is a $2^L D$ dimensional vector $\mathbf{x}^L$ that matches the dimensionality of $\mathbf{x}$. 

In each growing block, the input vector dimension is doubled in two ways, by direct upsampling and by a linear transform.  The resulting outputs are combined together and further transformed using a non-linear activation function:
\begin{equation}
    \mathbf{x}^{out}_{2d \times 1} =  q\left((1-\alpha)
    \begin{pmatrix}
    \mathbf{x}^{in}_{d \times 1}  \\
    \mathbf{x}^{in}_{d \times 1} 
    \end{pmatrix}
    +\alpha \ A_{2d \times d} \cdot \mathbf{x}^{in}_{d \times 1} \right)
    \label{eq:5}
\end{equation}
$A_{2d \times d}$ are trainable parameters in the linear transformation branch, $q(\cdot)$ is a non-linear activation function, and $\alpha$ is a hyperparameter that is manually tuned over the course of optimization.  

Initially, $\alpha$'s for all of the growing blocks in the network are set to $0$, such that the vector outputted by each block has the same effective dimensionality as its input vector.  The network output $\mathbf{x}^L$ therefore has an effective dimensionality that matches the dimensionality of the input $\mathbf{x}^0$. As $\alpha$ is increased for a particular growing block, its output vector becomes dominated by its linear transformation branch, as opposed to its upsampling branch, and it has an effective dimensionality that exceeds and eventually doubles that of the growing block input vector.  The effective dimensionality of $\mathbf{x}^L$ therefore arises from the aggregation of effective dimensionality increases from all growing blocks.  To control the effective dimensionality of $\mathbf{x}^L$ over the course of PG-GLOnet training, $\alpha$ is manually changed from 0 to 1 sequentially from the left to right blocks (bottom of Figure \ref{fig:1}b). At the end of PG-GLOnet training, $\alpha$  is $1$ for all growing blocks and the effective dimensionality of $\mathbf{x}^L$ matches that of $\mathbf{x}$.




\begin{figure*}[ht]
    \centering
    \includegraphics[width=\linewidth]{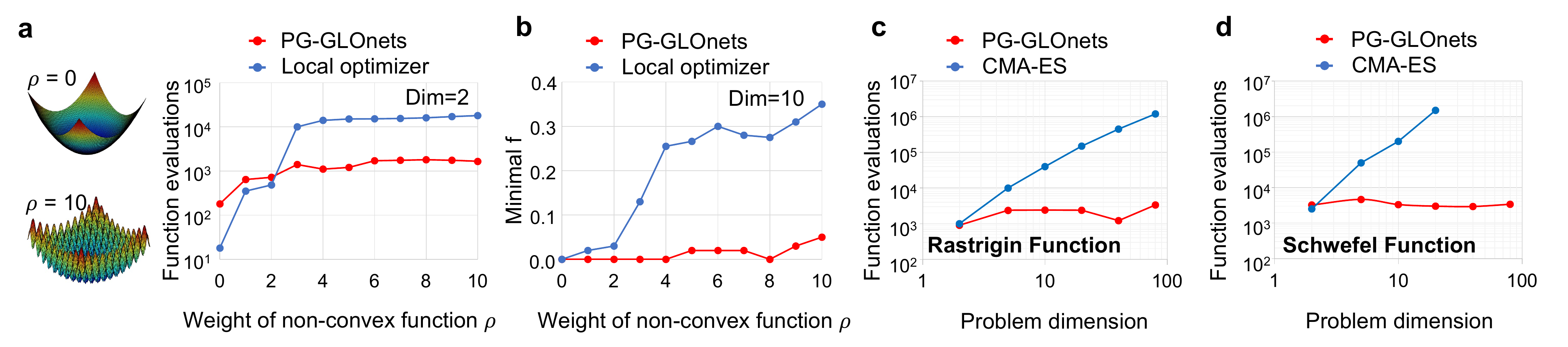}
    \caption{Benchmark results for PG-GLOnet. (a,b) Benchmark of PG-GLOnet with ADAM for a modified Rastrigin test function that can be tuned from convex to non-convex.  (a) Functional evaluations required by both algorithms for a two dimensional Rastrigin test function with differing $\rho$.  Plots of the test function for $\rho=0$ and $\rho=10$ are shown on the left. (b)  Optimal values achieved by PG-GLOnet and ADAM for a ten dimensional Rastrigin test function with differing $\rho$.  (c,d) Benchmark of PG-GLOnet with CMA-ES for modified (c) Rastrigin and (d) Schwefel test functions.  Both plots show the required number of functional evaluations required to find the global optimum as a function of test function dimension.}
    \label{fig:2}
\end{figure*}


To evaluate the efficacy of PG-GLOnet in solving high dimensional non-convex optimization problems, we perform a series of benchmark numerical experiments where we optimize a set of standard test functions with PG-GLOnet and other established algorithms.  In the first set of experiments, we consider a testing function that can be tuned from a convex to non-convex function and compare PG-GLOnet with ADAM, a well known momentum-based gradient descent algorithm that is typically more effective than gradient descent.  ADAM is a local optimization algorithm and performs well on convex objective functions but can get trapped within local optima for non-convex functions.  Our test function is a modified  Rastrigin function defined as follows:
\begin{equation}
    f(\mathbf{x}; \rho) = \rho d + \sum_{i=1}^d [x_i^2 - \rho\cos(2\pi x_i)]
    \label{eq:6}
\end{equation}
$\rho$ is a hyperparameter that specifies the amplitude of the sinusoidal modulation within the function.  When $\rho =0$, $f(\mathbf{x}; \rho) =  \sum_{i=1}^d x_i^2$  and is a convex function. As $\rho$ increases, more local optima emerge and these optima become separated by larger magnitude barriers.  

We first consider the computational cost required by ADAM and PG-GLOnet to find the global optimum of a two dimensional modified  Rastrigin function as a function of $\rho$.  For ADAM, we run 10000 optimizations for 200 iterations with random starting points, and for PG-GLOnet, we run the algorithm 10 times with a batch size of 20 for 200 total iterations. In both cases, the algorithms terminate early when they output results within $10^{-3}$ of the global optimum, and computational cost is quantified as the average number of function evaluations required  to find the global optimum.  The results are summarized in Figure \ref{fig:2}a and indicate that for convex or nearly convex optimization landscapes, ADAM is more efficient at finding the global optimum. This efficiency arises because ADAM is a specially tailored local optimizer that is well suited for these types of problems, while PG-GLOnet always requires relatively large batch sizes and more iterations to converge.  As $\rho$ increases, orders-of-magnitude more ADAM evaluations are required to search for the global optimum due to trapping within local optima in the design landscape.  The computational cost for PG-GLOnet, on the other hand, does not increase nearly as rapidly due to its ability to navigate non-convex landscapes and is ten times more efficient than ADAM for $\rho$ greater than $3$.


We also perform benchmarks between ADAM and PG-GLOnet for a ten dimensional problem.  Due to the inability for ADAM to converge to the global optimum in non-convex, high dimensional landscapes, we perform this benchmark differently and compare the best optimal value found by ADAM and PG-GLOnet given the same amount of computational resources. Here, we run ADAM for 200 iterations with 20 random starting points and PG-GLOnet for 200 iterations with a batch size of 20.  We run these benchmark experiments ten times and average the best values from each experiment, and the results are reported in Figure \ref{fig:2}b. 
 We find that the PG-GLOnet is able to consistently find solutions at or near the global optimum for all values of $\rho$, but the local optimizer gets progressively worse as $\rho$ increases.


In our next set of benchmark experiments, we compare PG-GLOnet with the covariance matrix adaptation evolution strategy (CMA-ES), which is an established evolutionary algorithm used to perform population-based global searching of an optimization landscape.  Compared to ADAM, it is more suitable for performing non-convex optimization.  We consider two standard non-convex testing functions with lots of local optima, the Rastrigin and Schwefel functions (defined in the Appendix). 

Plots in Figures \ref{fig:2}c and \ref{fig:2}d show the average number of function evaluations required to find the global optimum as a function of problem dimension $d$. The computational cost of CMA-ES increases exponentially as the problem dimension becomes larger, indicating the intractability of applying this algorithm to ultra-high dimensional problems.  For the Schwefel function, we limited our CMA-ES benchmarking experiments to a problem dimension of 20 due to this scaling trend.  PG-GLOnet, on the other hand, has a relatively small computational cost that is not sensitive to the dimension.  In fact, the same neural network architecture and batch size is used for all problems.  A more detailed discussion as to the origins of problem dimension and batch size decoupling is provided in the Discussion section.



Finally, we benchmark PG-GLOnet with state-of-art algorithms on testing functions proposed by the CEC'2013 Special Session and Competition on Large-Scale Global Optimization (LSGO) \cite{li2013benchmark}. We consider the six non-convex benchmark functions from the competition, which involve variations and combinations of the Rastrigin and Ackely functions and are defined in the Appendix.  These benchmark functions were designed to incorporate a number of challenging features for optimization, including:
\begin{enumerate}
    \item \textbf{High dimensions.} The design space of a optimization problem grows exponentially as the dimension of design variables increases. These benchmark functions utilize one thousand dimensional landscapes. 
    \item \textbf{Functions with non-separable subcomponents.} The whole design variable is decomposed into several subcomponents and dimensions within each subcomponent are strongly coupled together. 
    \item \textbf{Imbalance in the contribution of subcomponents.} The contribution of a subcomponent is magnified or dampened by a coefficient.  
    \item \textbf{Non-linear transformations to the base functions.} Three transformations are applied to break the symmetry and introduce some irregularity on the landscape: (1) Ill-conditioning (2) Irregularities (3) Symmetry breaking. 
\end{enumerate}

\begin{table*}[ht]
    \centering
    \caption{Optimization results from conjugate gradient, CC-RDG3, DGSC, FC-GLOnet followed by local gradient descent, and PG-GLOnet followed by local gradient descent, as applied to 1000-dimensional benchmark functions.}
    \begin{tabular}{c|p{3cm}|p{3cm}|p{3cm}|p{3cm}|p{3cm}}
    \hline
        ~ & CG & CC-RDG3 & DGSC & FC-GLOnet +Local Refinement & PG-GLOnet +Local Refinement\\ 
    \hline
        $f_1$ & (3.65 $\pm$ 0.07)e+04 & (2.36 $\pm$ 0.11)e+03 & \textbf{(7.15 $\pm$ 1.61)e+02} & (1.74 $\pm$ 0.01)e+03 & (1.85 $\pm$ 0.63)e+03\\ 
        $f_2$ & (1.98 $\pm$ 0.00)e+01 & (2.04 $\pm$ 0.00)e+01 & (2.07 $\pm$ 0.00)e+01 & (1.98 $\pm$ 0.00)e+01 &\textbf{(4.38 $\pm$ 0.75)e-03} \\ 
        $f_3$ & (2.08 $\pm$ 0.18)e+07 & (2.27 $\pm$ 0.30)e+06 & (3.27 $\pm$ 0.66)e+06 & (8.76 $\pm$ 2.99)e+05 &\textbf{(5.05 $\pm$ 0.86)e+05} \\ 
        $f_4$ & (9.76 $\pm$ 0.01)e+05 & (9.96 $\pm$ 0.00)e+05 & (1.06 $\pm$ 0.00)e+06 & (9.94 $\pm$ 0.01)e+05 &\textbf{(2.95 $\pm$ 0.82)e+02} \\ 
        $f_5$ & (1.15 $\pm$ 0.97)e+09 & (1.45 $\pm$ 0.32)e+08 & (1.79 $\pm$ 0.53)e+08 & \textbf{(4.22 $\pm$ 0.77)e+07} & (1.17 $\pm$ 0.49)e+08 \\ 
        $f_6$ & (8.84 $\pm$ 0.00)e+07 & (9.11 $\pm$ 0.14)e+07 & (9.38 $\pm$ 0.03)e+07 & (9.03 $\pm$ 0.01)e+07 &\textbf{(4.66 $\pm$ 0.79)e+04} \\ 
    \hline
    \end{tabular}
    \label{table:1}
\end{table*}

To globally search these landscapes for the global optimum, we perform a two step optimization procedure.  First, we run PG-GLOnet for each benchmark function for 200 iterations and a batch size of 100, from which our generative network outputs a narrow distribution of $\mathbf{x}$'s in promising regions of the optimization landscape.  We then sample this distribution 100 times and perform local gradient descent on each of these design variables for an additional 200 iterations.  The best function values found by PG-GLOnet plus local gradient descent are reported in Table \ref{table:1}, together with results produced from FC-GLOnet plus local gradient descent, local conjugate gradient descent, and two state-of-art non-convex optimization algorithms that were the best performing algorithms in the most recent LSGO contest: CC-RDG3, which is a divide-and-conquer method \cite{sun2019decomposition}, and DGSC, which is a differential group method utilizing spectral clustering \cite{li2019differential}. We observe that PG-GLOnet with local gradient descent refinement is able to significantly outperform the other algorithms for the majority of test functions. In addition, the total computational cost of the two step optimization procedure is only $4\times 10^4$ function evaluations, while CC-RDG3 and DGSC require $3\times 10^6$ function evaluations.

\section{Discussion}

We discuss the origins of the efficiency and efficacy of PG-GLOnet in solving ultra-high dimensional non-convex optimization problems.  First, we examine how the generic GLOnet algorithm operates and why it is able to effectively utilize a gradient-based strategy to solve non-convex optimization problems.  Second, we examine the role of the progressive growing generative network architecture in PG-GLOnet in solving ultra-high dimensional problems.  By understanding the relationship between network architecture and optimization procedure, we elucidate built-in assumptions used by PG-GLOnet in its search for the global optimum. 

\begin{figure*}[ht]
    \centering
    \includegraphics[width=15cm]{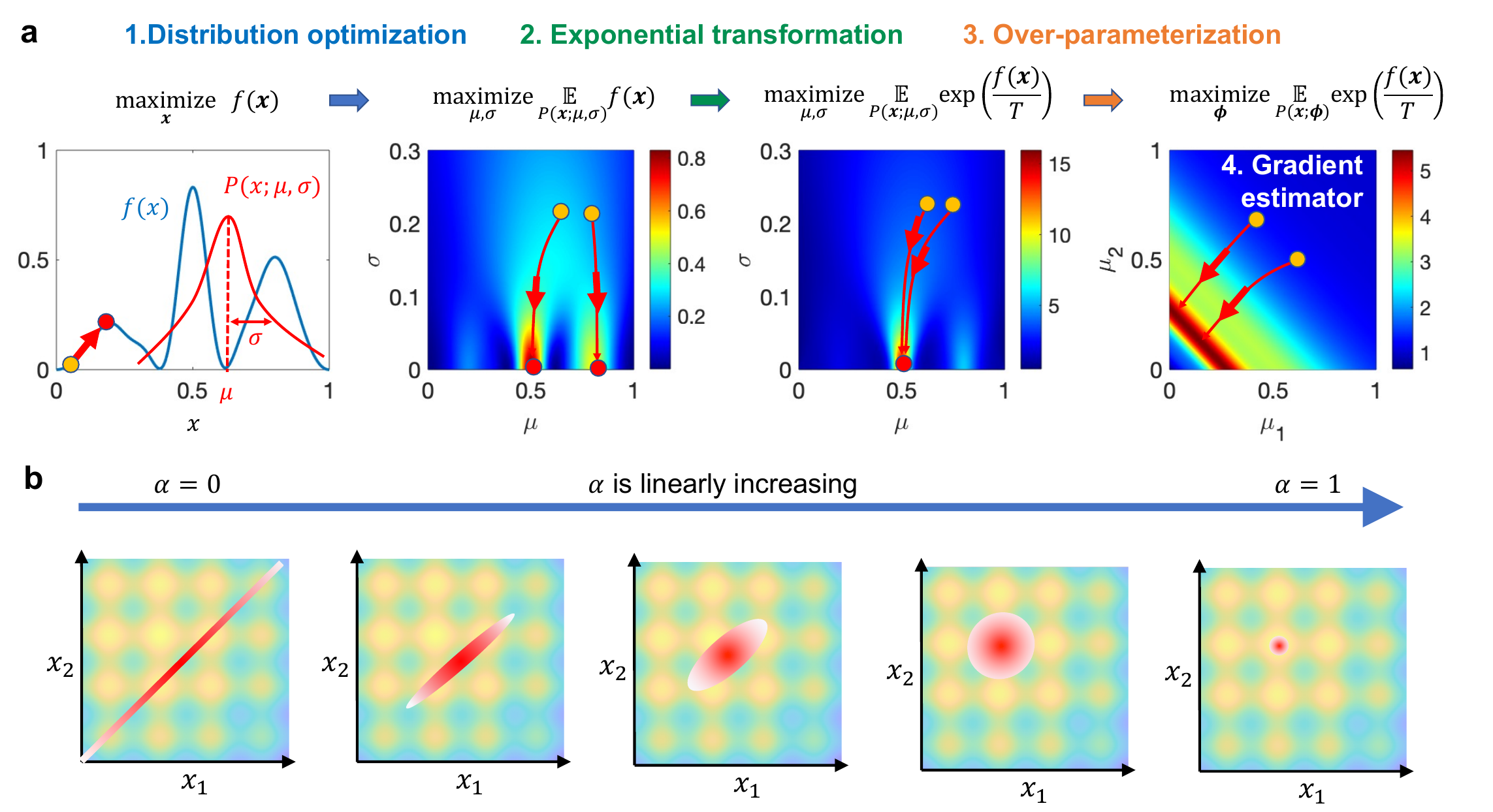}
    \caption{Conceptualization of the GLOnets optimization platform. (a) Visualization of key concepts that enable effective gradient-based optimization within a non-convex landscape, including: 1) transforming the optimization problem to the optimization of parameters within a distribution; 2) exponential weighing of the objective function; 3) over-parameterization of the distribution function; and 4) Effective gradient estimation during the network training procedure. b) Role of the progressively growing PG-GLOnet network architecture in optimization, visualized for a single growing block applied to a two-dimensional problem.  When $\alpha$ is zero, PG-GLOnet searches within a one dimensional slice of the two dimensional landscape.  As $\alpha$ increases, the effective dimensionality of the PG-GLOnet output distribution increases and enables searching of more of the landscape.  Upon the completion of PG-GLOnet training, the generator output distribution collapses to the global optimum.}
    \label{fig:3}
\end{figure*}

With the generic GLOnet algorithm, the original optimization problem cited in Equation 1 is reframed as a related problem (Equation 2) that addresses a transformed, smoothened optimization landscape.  The key concepts that produce this landscape transformation and enable effective gradient-based optimization are outlined in Figure \ref{fig:3}a and are: 1) distribution optimization, where the original problem involving the optimization of $\mathbf{x}$ is transformed to a problem involving the optimization of parameters within a simple distribution $P(\mathbf{x})$; 2) exponential transformation, where the objective function is exponentially weighted; 3) over-parametrization, where the distribution $P(\mathbf{x})$ is now parameterized by a neural network with hundreds to thousands of weights; and 4) gradient estimation, where gradients that specify the evolution of the continuous distribution $P(\mathbf{x})$ are accurately computed through discrete samplings of $\mathbf{z}$.

\textbf{Distribution optimization.}  With the concept of distribution optimization, the original problem of searching for an optimal $\mathbf{x}$ is recast as a population-based search in which parameters within a distribution function are optimized, thereby enabling a search for the global optimum in a smoother and higher dimensional optimization landscape.  This concept is shared by other population-based optimization algorithms, such as CMA-ES.  To visualize the concept, we consider a non-convex one-dimensional function $f(\mathbf{x})$ plotted as a blue line in the leftmost figure in Figure \ref{fig:3}a.  The objective is to maximize $f(\mathbf{x})$, and the function contains multiple local maxima separated by deep valleys.  It is easy for optimization algorithms, particularly gradient-based algorithms, to get trapped in the local optima.  For example, if gradient descent optimization is used and is initialized at the yellow dot position, the algorithm will converge to the local optimum delineated by the red dot.  With this approach, multiple independent gradient descent optimizations with random starting points are needed to increase the possibility of finding the global optimum. For these problems, gradient-free optimization heuristics are often employed, which can reduce the chances of trapping within suboptimal maxima but which introduce a more stochastic nature to the search process.

However, if we consider the optimization of a distribution function that interacts with the global optimization landscape, local information at different parts of the landscape can be aggregated and  collectively utilized to evolve this distribution in a manner that reduces issues of trapping within suboptimal maxima.  Formally, we transform the optimization variable $\mathbf{x}$ to parameters within the distribution $P(\mathbf{x})$, and the globally optimal distribution is one that is narrowly peaked around the global optimum.  Distribution functions can be explicitly parameterized in many ways.  As a simple illustrative example that builds on our 
discussion of the one-dimensional $f(\mathbf{x})$, we consider the one-dimensional Gaussian distribution denoted as $P(\mathbf{x; \mu, \sigma})$, shown as the red curve in the leftmost figure in Figure \ref{fig:3}a. $\mu$ and $\sigma$ refer to mean and standard deviation, respectively.


With a Gaussian distribution function, the objective function now becomes transformed to the expected value of $f(\mathbf{x})$ as a function of $(\mu, \sigma)$: $\mathop{\mathbb{E}}_{\mathbf{x} \sim P(\mathbf{x}; \mu, \sigma)} f(\mathbf{x})$. As this new optimization landscape is a function of two distribution parameters, $\mu$ and $\sigma$, it is two dimensional.  We can directly visualize this new landscape by evaluating $\int f(\mathbf{x}) P(\mathbf{x};\mu, \sigma) d\mathbf{x}$ for all values of $(\mu, \sigma)$, and the result is summarized in the second figure from the left in Figure \ref{fig:3}a.  The horizontal line section at the bottom of the contour plot, where $\sigma$ equals zero, is the  original one-dimensional $f(\mathbf{x})$ with multiple optima.  As $\sigma$ increases to finite values above zero, the landscape becomes smoother.  Mathematically, horizontal line sections for finite sigma are calculated by convolving $f(\mathbf{x})$ with the Gaussian function, producing a Gaussian blur that leads to smoothening.  This smoothened landscape facilitates gradient-based optimization of $(\mu, \sigma)$ when the distribution is initialized to large $\sigma$ values, and the final optimized distributions converge to the original $f(\mathbf{x})$ space at the bottom of the plot.  However, while this two-dimensional landscape is smoother than the original $f(\mathbf{x})$, there remain multiple distribution parameter initializations for which the gradient-based optimizer converges to suboptimal maxima.







\textbf{Exponential transformation.} To further smoothen the optimization landscape and enhance the presence of the global optimum, we perform an exponential transformation of the objective function.  Mathematically, the objective function for the distribution optimization problem becomes: $\mathop{\mathbb{E}}_{\mathbf{x} \sim P(\mathbf{x}; \mu, \sigma)} \exp\left[ \frac{f(\mathbf{x})}{T} \right]$. The temperature term $T$ modulates the impact of the global optimum on the optimization landscape such that low $T$ produces strong landscape modulation by the global optimum.  For our one-dimensional $f(\mathbf{x})$ example, the exponentially transformed landscape is plotted in the second figure from the left in Figure \ref{fig:3}a and shows that the local optima has faded out, such that gradient-based optimization within this landscape is more likely to converge to the global optimum.

The choice of $T$ depends on the scale of $f(\mathbf{x})$. Consider $f(\mathbf{x})$ that is linearly normalized to span $(0, 1)$.  Such normalization can be typically achieved based on prior knowledge about the upper and lower bound of $f(\mathbf{x})$.  If we want to amplify $f(\mathbf{x})$ for $f(\mathbf{x}) > f_d$ and minimize $f(\mathbf{x})$ for $f(\mathbf{x}) < f_d$, where $f_d$ is a division point between 0 and 1, the temperature is chosen to be $T = f_d / \log(1 + f_d)$. For example, if $f_d$ is chosen to be the golden ratio, then the temperature is roughly $T = 1.3$. In practice, the selection of $f_d$ is problem specific, and $T$ can be treated as a hyperparameter that can be manually tuned around 1 for tailoring to a particular problem. 


\textbf{Over-parameterization.} To further enhance the ability for GLOnet to efficiently and reliably converge to the global optimum, we next consider the concept of over-parameterization in which the distribution $P(\mathbf{x})$ is now a neural network parameterized by weights $\mathbf{\phi}$.  The objective function then becomes: $\mathop{\mathbb{E}}_{\mathbf{x} \sim P(\mathbf{x}; \phi)} \exp\left[ \frac{f(\mathbf{x})}{T} \right]$. Our use of a neural network is inspired by the fact that deep network training involves the solving of an extremely high dimensional non-convex optimization problem, that the convergence of the neural network is typically insensitive to initialization, and that good neural network parameters can be found using backpropagation. 

The underlying mathematical principles outlining why gradient descent is so effective for  deep network training have been revealed to some extent by computer scientists in recent years. \cite{choromanska2015loss, arora2018optimization, draxler2018essentially}  First, the parameter space of deep networks is a high-dimensional manifold, such that most local optima are equivalently good and the probability of converging to a bad optimum during training decreases quickly with network size.  Second, these equivalently high performing local optima originate from neural network over-parameterization, which builds in redundancy in the optimization landscape that speeds up and stabilizes the gradient-based optimization process. 

To understand how this applies to GLOnet, we revisit our one-dimensional $f(\mathbf{x})$ landscape in which local optima are separated by deep barriers.  When the optimization landscape is transformed using $P(\mathbf{x,\phi})$, it frames the optimization problem in a very high dimensional landscape, as the dimensionality of $\mathbf{\phi}$ is much higher than $\mathbf{x}$.  Solutions to the optimization problem therefore reside in a high-dimensional manifold, such that many different $\mathbf{\phi}$'s serve as high performing local optima.  Additionally, local optima in $f(\mathbf{x})$ are no longer separated by deep barriers but are instead connected by pathways with low to no barriers in our transformed high dimensional landscape, mitigating trapping within these local optima during gradient-based optimization. The high dimensional landscape representing the transformed $f(\mathbf{x})$ is visualized as a two-dimensional projection in the rightmost plot in Figure \ref{fig:3}a.  The global optimum is now a connected band in the optimization landscape, as opposed to a single point in $f(\mathbf{x})$, and there are fewer energy barriers preventing gradients from converging to the global optimum, enabling gradient descent optimization to be more robust and faster. We note that neural network depth and expressivity play a large role in determining the practical impact of over-parameterization on optimization, and as a demonstration, we compare the performance of GLOnets based on linear and deep non-linear networks in the Appendix.

\textbf{Gradient estimation.} A critical feature to maximizing the performance of GLOnet is ensuring that gradients used to evolve $P(\mathbf{x})$, which are approximated using a finite batch of samples, are sufficiently accurate.  There are two methods for gradient estimation that can be used for GLOnets. The first is to use a score function gradient estimator, which utilizes the evaluated derivatives of the probability distribution $P(\mathbf{x}; \phi)$ and $f(\mathbf{x})$. This method for estimation requires explicit evaluation of derivatives to $P(\mathbf{x}; \phi)$ but only an implicit evaluation of $\nabla_{\mathbf{x}}f$.  The second is to use a pathwise gradient estimator, which relies on knowing the explicit derivatives of $f(\mathbf{x})$ but for which the probability distribution $P(\mathbf{x}; \phi)$ can be implicit. Empirically, we find for GLOnet that the pathwise gradient estimator more consistently produces smaller gradient error compared with the score function gradient estimator, and we therefore implement the pathwise gradient estimator in Equation \ref{eq:4}. \cite{xu2019variance, mohamed2020monte} 

The pathwise gradient estimator is based on the principle of Monte Carlo estimation, such that the estimation error decreases with the inverse square root of batch size.  Importantly, this estimation error is independent of dimension. As a result, GLOnet and specifically PG-GLOnet are able to operate for batch sizes that are independent of problem dimension, as demonstrated in Figures 2c and 2d.  This scaling of problem dimension without a required scaling in the number of functional evaluations allows PG-GLOnet to readily scale and address the 1000-dimensional problems in Table 1 with modest computational resources.






\textbf{Progressive growth.}  Direct searching within a high dimensional, non-convex landscape is an intractable problem.  In the case of FC-GLOnet, which utilizes all of the features above, including distribution optimization and over-parameterization, the algorithm is still not effective in directly searching high dimensional landscapes (Table 1).  With PG-GLOnet, the progressive growing architecture regularizes the optimization procedure to search first within a relatively coarse, low dimensional representation of the optimization landscape, followed by relatively local searching within increasingly higher dimensional landscape representations.  This hierarchical increase of landscape dimensionality directly corresponds to the serial toggling of $\alpha$ within the series of growing blocks in the generator.  As such, the optimization landscape is evolved over the course of PG-GLOnet training in a manner that  maintains the tractability of the optimization problem.

To further visualize the relationship between generative network architecture and optimization search procedure, we consider a non-convex two-dimensional landscape shown in Figure \ref{fig:3}b.  The generative network contains a single growing block, and the toggling of $\alpha$ from zero to one modulates the effective dimensionality of the generator output from one to two.  Initially, $\alpha$ is zero and the vector outputted by the generator has the same effective dimensionality as its input vector and is one.  The optimization landscape being searched is therefore a diagonal line within the  two-dimensional landscape (Figure \ref{fig:3}b, left-most plot), and with optimal solutions near the center of the line, the outputted generator distribution (red coloring in plot) narrows towards this region.  As $\alpha$ is increased, the generator output vector becomes dominated by its linear transformation branch, as opposed to its upsampling branch, and it has an effective dimensionality that increases and eventually doubles.  In our PG-GLOnet visualization, this increase in effective dimensionality corresponds to a broadening of the optimization landscape being searched, and the outputted generator distribution widens relative to the diagonal line.  Upon the completion of network growth, the PG-GLOnet distribution converges to the global optimum.

The success of PG-GLOnet is therefore predicated on the ability for the outputted distribution of the generative network to be narrowed down to smaller but more promising regions of a coarse optimization landscape, prior to increasing the landscape dimensionality and adding more degrees of freedom to the problem.  This concept therefore works particularly well for problems where optima within a low dimensional analogue of the optimization landscape help to inform of the presence and position of optima within the high dimensional landscape.  This regularization of the optimization procedure also indicates that for problems where optima within coarse variants of the optimization landscape do not inform the position of the global optimum, PG-GLOnet will not work well.


In summary, we present a general global optimization algorithm metaheuristic based on progressive growing deep generative neural networks termed PG-GLOnet. Unlike other population-based algorithms,  PG-GLOnet uses gradient-based optimization to evolve an expressive, complex distribution in the optimization landscape to one centered around promising optima.  This complex distribution, parameterized using the deep network framework, utilizes loss function engineering and over-parameterization to facilitate effective gradient-based searching.  PG-GLOnet is particularly well suited to address ultra-high dimensional problems because the required batch size is independent of problem dimension and the progressively growing network architecture facilitates a hierarchical search process within a landscape with progressively growing effective dimensionality.  This use of a hierarchical search strategy also provides bounds as to the types of problems and landscapes that are suited for PG-GLOnet optimization.  We anticipate that further research in the tailoring of application-specific generative network architectures to particular optimization landscapes will enable the GLOnet platform to extend and adapt to an even wider range of non-convex, high dimensional optimization problems.


\bibliography{refs} 

\begin{thebibliography}{20}
\providecommand{\natexlab}[1]{#1}

\bibitem[{Arora, Cohen, and Hazan(2018)}]{arora2018optimization}
Arora, S.; Cohen, N.; and Hazan, E. 2018.
\newblock On the optimization of deep networks: Implicit acceleration by
  overparameterization.
\newblock In \emph{International Conference on Machine Learning}, 244--253.
  PMLR.

\bibitem[{Balsa-Canto et~al.(2012)Balsa-Canto, Banga, Egea,
  Fernandez-Villaverde, and de~Hijas-Liste}]{balsa2012global}
Balsa-Canto, E.; Banga, J.~R.; Egea, J.~A.; Fernandez-Villaverde, A.; and
  de~Hijas-Liste, G. 2012.
\newblock Global optimization in systems biology: stochastic methods and their
  applications.
\newblock In \emph{Advances in Systems Biology}, 409--424. Springer.

\bibitem[{Chen et~al.(2017)Chen, Jia, Jiang, Li, and Wang}]{chen2017sgo}
Chen, Z.; Jia, W.; Jiang, X.; Li, S.-S.; and Wang, L.-W. 2017.
\newblock SGO: A fast engine for ab initio atomic structure global optimization
  by differential evolution.
\newblock \emph{Computer Physics Communications}, 219: 35--44.

\bibitem[{Choromanska et~al.(2015)Choromanska, Henaff, Mathieu, Arous, and
  LeCun}]{choromanska2015loss}
Choromanska, A.; Henaff, M.; Mathieu, M.; Arous, G.~B.; and LeCun, Y. 2015.
\newblock The loss surfaces of multilayer networks.
\newblock In \emph{Artificial intelligence and statistics}, 192--204. PMLR.

\bibitem[{Draxler et~al.(2018)Draxler, Veschgini, Salmhofer, and
  Hamprecht}]{draxler2018essentially}
Draxler, F.; Veschgini, K.; Salmhofer, M.; and Hamprecht, F. 2018.
\newblock Essentially no barriers in neural network energy landscape.
\newblock In \emph{International conference on machine learning}, 1309--1318.
  PMLR.

\bibitem[{Held et~al.(2011)Held, Korte, Rautenbach, and
  Vygen}]{held2011combinatorial}
Held, S.; Korte, B.; Rautenbach, D.; and Vygen, J. 2011.
\newblock Combinatorial optimization in VLSI design.
\newblock \emph{Combinatorial Optimization}, 33--96.

\bibitem[{Jiang and Fan(2019)}]{jiang2019global}
Jiang, J.; and Fan, J.~A. 2019.
\newblock Global optimization of dielectric metasurfaces using a physics-driven
  neural network.
\newblock \emph{Nano letters}, 19(8): 5366--5372.

\bibitem[{Jiang and Fan(2020)}]{jiang2020simulator}
Jiang, J.; and Fan, J.~A. 2020.
\newblock Simulator-based training of generative neural networks for the
  inverse design of metasurfaces.
\newblock \emph{Nanophotonics}, 9(5): 1059--1069.

\bibitem[{J{\o}rgensen et~al.(2018)J{\o}rgensen, Larsen, Jacobsen, and
  Hammer}]{jorgensen2018exploration}
J{\o}rgensen, M.~S.; Larsen, U.~F.; Jacobsen, K.~W.; and Hammer, B. 2018.
\newblock Exploration versus exploitation in global atomistic structure
  optimization.
\newblock \emph{The Journal of Physical Chemistry A}, 122(5): 1504--1509.

\bibitem[{Karras et~al.(2017)Karras, Aila, Laine, and
  Lehtinen}]{karras2017progressive}
Karras, T.; Aila, T.; Laine, S.; and Lehtinen, J. 2017.
\newblock Progressive growing of gans for improved quality, stability, and
  variation.
\newblock \emph{arXiv preprint arXiv:1710.10196}.

\bibitem[{Li et~al.(2019)Li, Fang, Wang, and Sun}]{li2019differential}
Li, L.; Fang, W.; Wang, Q.; and Sun, J. 2019.
\newblock Differential grouping with spectral clustering for large scale global
  optimization.
\newblock In \emph{2019 IEEE Congress on Evolutionary Computation (CEC)},
  334--341. IEEE.

\bibitem[{Li et~al.(2013)Li, Tang, Omidvar, Yang, Qin, and
  China}]{li2013benchmark}
Li, X.; Tang, K.; Omidvar, M.~N.; Yang, Z.; Qin, K.; and China, H. 2013.
\newblock Benchmark functions for the CEC 2013 special session and competition
  on large-scale global optimization.
\newblock \emph{gene}, 7(33): 8.

\bibitem[{Mohamed et~al.(2020)Mohamed, Rosca, Figurnov, and
  Mnih}]{mohamed2020monte}
Mohamed, S.; Rosca, M.; Figurnov, M.; and Mnih, A. 2020.
\newblock Monte Carlo Gradient Estimation in Machine Learning.
\newblock \emph{J. Mach. Learn. Res.}, 21(132): 1--62.

\bibitem[{Pelikan et~al.(1999)Pelikan, Goldberg, Cant{\'u}-Paz
  et~al.}]{pelikan1999boa}
Pelikan, M.; Goldberg, D.~E.; Cant{\'u}-Paz, E.; et~al. 1999.
\newblock BOA: The Bayesian optimization algorithm.
\newblock In \emph{Proceedings of the genetic and evolutionary computation
  conference GECCO-99}, volume~1, 525--532. Citeseer.

\bibitem[{Piggott et~al.(2015)Piggott, Lu, Lagoudakis, Petykiewicz, Babinec,
  and Vu{\v{c}}kovi{\'c}}]{piggott2015inverse}
Piggott, A.~Y.; Lu, J.; Lagoudakis, K.~G.; Petykiewicz, J.; Babinec, T.~M.; and
  Vu{\v{c}}kovi{\'c}, J. 2015.
\newblock Inverse design and demonstration of a compact and broadband on-chip
  wavelength demultiplexer.
\newblock \emph{Nature Photonics}, 9(6): 374--377.

\bibitem[{Sell et~al.(2017)Sell, Yang, Doshay, Yang, and Fan}]{sell2017large}
Sell, D.; Yang, J.; Doshay, S.; Yang, R.; and Fan, J.~A. 2017.
\newblock Large-angle, multifunctional metagratings based on freeform multimode
  geometries.
\newblock \emph{Nano letters}, 17(6): 3752--3757.

\bibitem[{Sieniutycz and Jezowski(2009)}]{sieniutycz2009energy}
Sieniutycz, S.; and Jezowski, J. 2009.
\newblock \emph{Energy optimization in process systems}.
\newblock Elsevier.

\bibitem[{Snoek et~al.(2015)Snoek, Rippel, Swersky, Kiros, Satish, Sundaram,
  Patwary, Prabhat, and Adams}]{snoek2015scalable}
Snoek, J.; Rippel, O.; Swersky, K.; Kiros, R.; Satish, N.; Sundaram, N.;
  Patwary, M.; Prabhat, M.; and Adams, R. 2015.
\newblock Scalable bayesian optimization using deep neural networks.
\newblock In \emph{International conference on machine learning}, 2171--2180.
  PMLR.

\bibitem[{Sun et~al.(2019)Sun, Li, Ernst, and Omidvar}]{sun2019decomposition}
Sun, Y.; Li, X.; Ernst, A.; and Omidvar, M.~N. 2019.
\newblock Decomposition for large-scale optimization problems with overlapping
  components.
\newblock In \emph{2019 IEEE congress on evolutionary computation (CEC)},
  326--333. IEEE.

\bibitem[{Xu et~al.(2019)Xu, Quiroz, Kohn, and Sisson}]{xu2019variance}
Xu, M.; Quiroz, M.; Kohn, R.; and Sisson, S.~A. 2019.
\newblock Variance reduction properties of the reparameterization trick.
\newblock In \emph{The 22nd International Conference on Artificial Intelligence
  and Statistics}, 2711--2720. PMLR.

\end{thebibliography}

\end{document}